# Original Paper

Papis Wongchaisuwat, MS[1], Diego Klabjan, PhD[1]; Siddhartha R. Jonnalagadda, PhD[2];

[1]Department of Industrial Engineering and Management Sciences, Northwestern University, Evanston, IL, ²Divison of Health and Biomedical Informatics, Feinberg School of Medicine, Northwestern University, Chicago, IL.

**Title: A Semi-supervised learning approach to enhance health care Community-based Question Answering: A case study in alcoholism**

## ABSTRACT

**Background.** Community-based Question Answering (CQA) sites play an important role in addressing health information needs. However, a significant number of posted questions remain unanswered. Automatically answering the posted questions can provide a useful source of information for online health communities.

**Objective.** In this study, we developed an algorithm to automatically answer health-related questions based on past questions and answers (QA). We also aimed to understand information embedded within online health content that are good features in identifying valid answers.

**Methods.** Our proposed algorithm uses information retrieval techniques to identify candidate answers from resolved QA. In order to rank these candidates, we implemented a semi-supervised leaning algorithm that extracts the best answer to a question. We assessed this approach on a curated corpus from Yahoo! Answers and compared against a rule-based string similarity baseline.

**Results.** On our dataset, the semi-supervised learning algorithm has an accuracy of 86.2%. UMLS-based (health-related) features used in the model enhance the algorithm's performance by proximately 8 %. A reasonably high rate of accuracy is obtained given that the data is considerably noisy. Important features distinguishing a valid answer from an invalid answer include text length, number of stop words contained in a test question, a distance between the test question and other questions in the corpus as well as a number of overlapping health-related terms between questions.

**Conclusions.** Overall, our automated QA system based on historical QA pairs is shown to be effective according to the data set in this case study. It is developed for general use in the health care domain which can also be applied to other CQA sites.

**Keywords.** machine learning; natural language processing; question answering; online health communities; consumer health informatics

## INTRODUCTION

A study by Pew Internet Project's research reported 87% of U.S. adults use the Internet, and 72% of Internet users sought health information online in the past year [1]. Other studies have also analyzed the modes in which health information is shared and its impact on consumer decision making [2, 3]. While it is known that patients are seeking information that might not be obtained during the course of their regular clinical care and valuable knowledge is publicly available online, it is not trivial for users to quickly find an accurate answer to specific questions. Consequently, Community-based Question Answering (CQA) sites such as Yahoo! Answers tend to be a potential solution to this challenge. In CQA sites, users post a question and expect the online health community to promptly provide desirable answers. Despite a high volume of users' participation, a considerable number of questions are left unanswered and at the same time other questions that address the same information need are answered elsewhere. This common situation drew our attention to develop an automated system for answering both unsuccessfully answered and newly posted questions.

Substantial research exists for developing systems that address physicians' information needs at the point-of-care. Infobuttons and other decision support tools automatically select and retrieve information from knowledge sources at the point-of-care [4]. Social media platforms involve exchanges of health information among peers at any place and time [5]. The advantages and disadvantages of using a social network to address the information needs compared with a search engine are described in [6]. However, limited research has been done in addressing the information needs of patients through automated approaches that synthesize the information shared across online health communities. CQA systems in the health care domain address this issue.

QA systems are widely studied in both open and other restricted domains. One of the common approaches is to retrieve answers based on past QA, which is also fundamental to our work. Shtok et al. [7] extracted an answer from resolved QA pairs obtained from Yahoo! Answers. Specifically, a statistical model was implemented to estimate the probability that the best answer from the past posts can satisfactorily answer a newly posted question. In addition to Shtok et al., Marom et al. [8] implemented a predictive model involving a decision graph to generate help-desk responses from historical email dialogues between users and help-desk operators. Feng et al. [9] constructed a system aiming to provide accurate responses to students' discussion board questions. An important element in these QA systems is identifying the closest (the most similar) matching between a new question and other questions in a corpus. However, this is not a trivial task since both the syntactic and semantic structure of sentences should be considered in order to achieve an accurate matching. A syntactic tree matching approach was proposed to tackle this problem in CQA [10]. Jeon et al. [11] developed a translation-based retrieval model exploiting word relationships to determine similar questions in QA archives. Various string similarity measures were also implemented to directly compute the distance between two different strings [12]. A topic clustering approach was introduced to find similar questions among QA pairs [13].

An important component in QA systems is re-ranking of candidates in order to identify the best answer. A probabilistic answer selection framework was used to estimate the probability of an answer candidate being correct [14]. Alternatively, supervised learning-based approaches including support vector machine [15, 16] and logistic regression [17] are applicable to select (rank) answers. Commonly, collecting a large number of labeled data can be very expensive or even impossible in practice. Wu et al. [18] developed a novel unsupervised support vector machine classifier to overcome this problem. Other studies used different classifiers with multiple features for similar problems [19-23].

Athenikos et al. [24] conducted a thorough survey reviewing state of the art in biomedical question answering systems. Morris et al. [25] presented a survey study about the behavior of users in question and answer systems. Luo et al. [26] developed an algorithm, SimQ, to extract similar consumer health questions based on both syntactic and semantic analysis. Vector-based distance measures were used to compute similarity score among questions. Statistical syntactic parsing and standardized unified medical language system (UMLS) were implemented to construct syntactic and semantic features, respectively. However, to effectively use the information in CQAs, we need to not only retrieve similar questions, but also provide and validate potential answers. SimQ was designed to retrieve similar questions from the NetWellness [27], a health information platform that has been maintained by clinician peer-reviewers. Questions collected within NetWellness tend to be clean and well structured, while CQA websites tend to be noisy. Wong et al. has also contributed to automatically answering health-related questions based on previously solved QA pairs [28]. They provide an interactive system where the input questions are precise and short as opposed to accepting CQA questions directly as input.

In comparison to these systems, our work relies on implementing semi-supervised learning with Expectation Maximization (EM) approach [29]. Semi-supervised learning uses both labeled and unlabeled data for training. Given labeled and unlabeled data, EM based semi-supervised learning first trains an initial model using just the labeled set. This model is then used to estimate the label of each element in the unlabeled set. Next, the model is re-trained using both labeled and unlabeled set with the estimated labels from the previous step. The new model is used to refine the estimated labels in the unlabeled set. These steps are iteratively repeated until the algorithm converges or reaches pre-defined number of iterations. In addition, we employed Dynamic Time Warping [30] along with the vector-space distance [31] to measure similarity and incorporated biomedical concepts as additional features.

In summary, our work aims to automatically answer health-related questions based on past QA. We extracted candidate questions based on similarity measure and selected possible answers by using a semi-supervised learning algorithm. Automatically retrieving answers for questions from online health communities should provide the users a potential source of health information.

## METHODS

The system was built as a pipeline that involves two phases. The first phase implemented as a rule-based system, consists of: A) *Question Extracting*, which maps the Yahoo! Answers dataset to a data structure that includes question category, the short version of the question and the two best answers; B) *Answer Extracting*, which employs similarity measures to find answers for a question from existing QA pairs. In the second phase of *Answer Re-ranking*, we implemented supervised and semi-supervised learning models that refined the output of the first phase by screening out invalid answers and ranking the remaining valid answers.

Figure 1 depicts the system architecture and flow. In training, phase I is applied for each prospective question in the training data set (with all other questions under a consideration corresponding to all questions in the corpus being different from the current prospective question). For test, the prospective question is a test question and all other questions are those from the training set. In this case, phase II uses the trained model to rank the candidate answer.

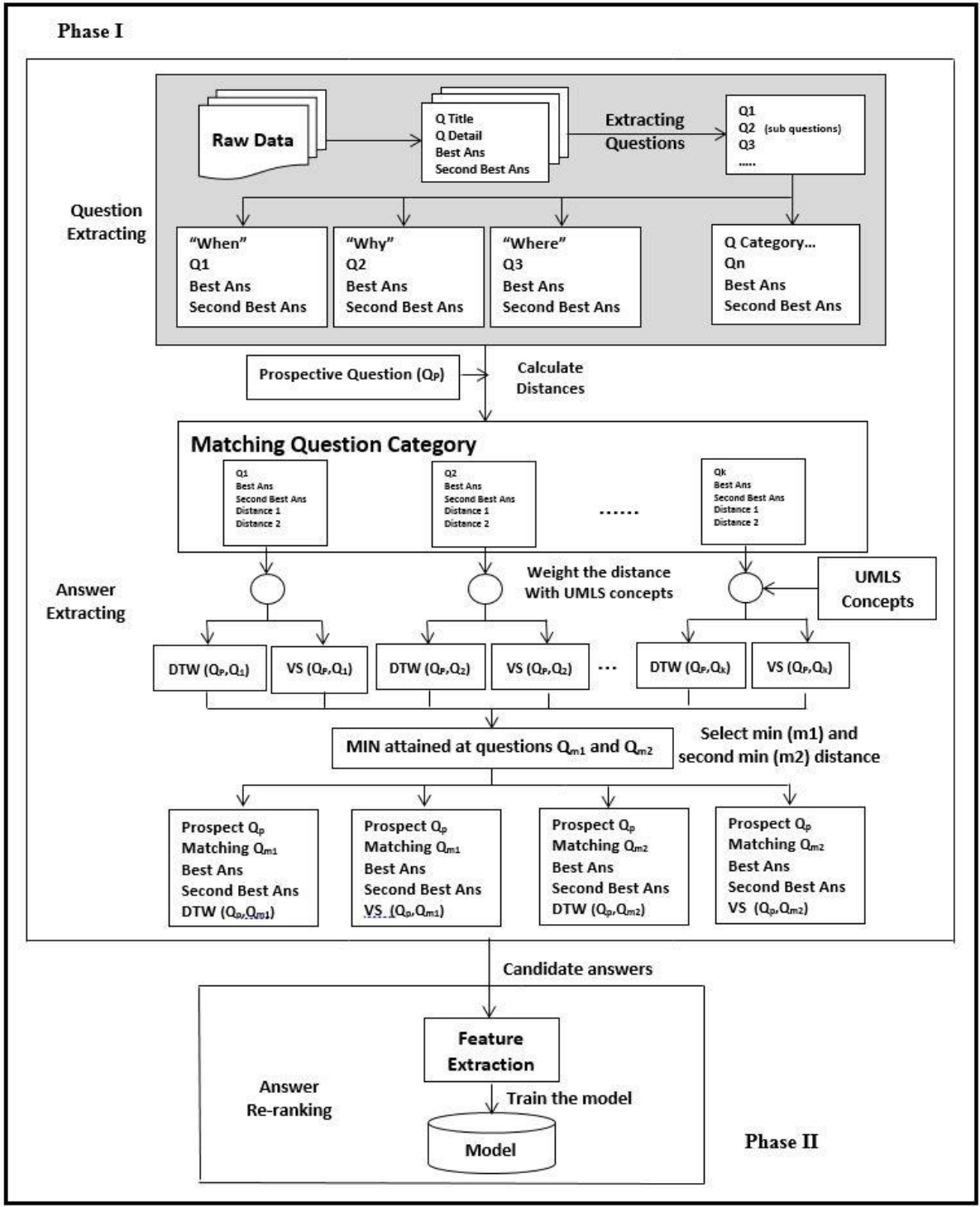

**Figure 1: Overall architecture for training the system.**

We first describe the training phase. The Rule-based Answer Extraction phase (phase I) is split into 2 steps.

A) Question Extracting: For this system, we assumed that each question posted on CQA sites has a question title and its description. Once users provided possible answers to the posted question, some responses were assumed to be marked as the best answer either by the question provider or community users. The second and subsequent best answers were chosen among remaining answers based on the number of likes. The raw data collected from CQA sites is unstructured and contains unnecessary text. It is essential to retrieve short and precise questions embedded in the original question title and its description (which can include up to 4-5 question sentences). Instead of using the whole question title and description which are long and verbose, we implemented a rule-based approach to capture these possible short question sentences (sub-questions). These sub-questions were categorized into different groups based on the words in questions. More specifically, regular expressions based on question words were used to classify sub-questions, which yielded different question classes consisting of "Yes-No," "what quantity," "how frequent," "when," "why," "how," "where," "who," "whose," "whom," "what," and "which," and "others." We considered sub-questions, instead of full questions and descriptions, in the rest of the paper.

B) Answer Extracting: Given a question, it was divided into sub-questions and matched with the question group using the above rule-based approach. Then, we computed the semantic distance between the prospective question and all other questions from the training sets belonging to the same group. Two distance approaches were employed in our work.

1. DTW-based approach: It is based on a sequence alignment algorithm known as Dynamic Time Warping [30], which employs efficient dynamic programming to calculate a distance between two temporal sequences. This allows us to effectively encode the word order without adversely penalizing for missing words (such as in a relative clause). Applying it in our context, a sentence was considered as a sequence of words where the distance between each word was computed by the Levenshtein distance at a character level [32, 33]. For any two sequences defined as $Seq_1 = <w_1^1, w_2^1, \ldots, w_m^1>$ and $Seq_2 = <w_1^2, w_2^2, \ldots, w_n^2>$ where m and n are the lengths of the sequences, Liu et al [30] defined the distance between two sequences (in our case, two sentences) as below:

$$D_{seq}(seq_1, seq_2) = f(m,n) \quad \text{and} \quad f(i,j) = d(w_i^1, w_j^2) + \min \begin{cases} f(i-1,j) \\ f(i,j-1) \\ f(i-1,j-1) \end{cases}$$

where $f(0,0) = 0, f(i,0) = f(0,j) = \infty, i \in (0,m), j \in (0,n)$

Here $d(w_i^1, w_j^2)$ is the distance between two words computed by the Levenshtein measure.

2. Vector-space based approach: An alternative paradigm is to consider the sentences as a bag of words, represent them as points in a multi-dimensional space of individual words and then calculate the distance between them. We implemented a unigram model with tf-idf weights based on the prospective question and other questions in the same category and computed the Euclidean distance measure.

We further took into account the cases that share similar medical information by multiplying the distances with a given weight parameter. The best value of the weight parameter was selected based on extensive experiments. The MetaMap tool was used to recognize UMLS concepts occurring in questions [34]. If at least one word in the UMLS concepts of "organic chemical" and "pharmacologic substance" occurs in both the prospective question and a training question, we reduce the distance to account for the additional semantic similarity. These UMLS concepts are specifically selected as we want to provide more weight to answers that mention a treatment approach under the intuitive assumption that the majority of CQA users aim to seek informative advice for their illness. The set of semantic types can be expanded to capture broader concepts if different domains are considered.

The QA pairs in the training set corresponding to the smallest and the second smallest distance were extracted. Thus, we finally obtained a list of candidate answers, i.e. the answers referring to smallest and second smallest questions, for each prospective question. These answers were used as the output of the baseline rule-based system. This was repeated for each question in the training set, i.e. the prospective question corresponds to each question in the training set. At the end of this phase we had triplets ($Q_p$, $Q_t$, $A_t$) over all questions $Q_p$. Note that $A_t$ is an answer to question $Q_t$ with $Q_t \neq Q_p$ and each $Q_p$ yielded several such triplets.

The machine learning phase of answer re-ranking (phase II) is described next. The goal of this phase is to rank candidate answers from the previous step and select the best answer among them. Each triple ($Q_p$, $Q_t$, $A_t$) is aimed to be assigned as "valid" if $A_t$ is a valid answer to $Q_p$, or "invalid" otherwise. We describe how the model was trained in this section while detailed explanations (e.g. number of labeled and unlabeled triplets) are provided in the results section. We first selected a small random subset of triplets and labeled them manually (there were too many to label all of them in this way). Both supervised and semi-supervised learning Expectation Maximization (EM) models were developed to predict the answerability of newly posted question as well as rank candidate answers. Specifically, the semi-supervised learning model was trained on labeled and unlabeled triplets. According to the semi-supervised learning model, we first trained a supervised learning algorithm including Neural Networks with the entropy objective function (NNET), Neural Networks with the L2-norm or least squares objective function (NNET_L2), Support Vector Machine (SVM), and Logistic Regression (LOG) based on manually labeling outputs from the above rule-based answer extraction phase. The trained model was used to classify the unlabeled part of the outputs of phase I, and then the classifier was retrained based on the original labeled data and a randomly selected subset of unlabeled data using the estimated labels from the previous iteration. These steps were iteratively repeated in order to achieve a final estimated label. The supervised approach, on the other hand, only ran a classifier on the labeled subset and finished. A 10-fold cross validation was implemented in both semi-supervised and supervised approaches. Specifically, all labeled observations were partitioned into 10 parts where one part was set aside as a test set. The model was fitted based on the remaining 9 parts of the labeled observations (plus the entire unlabeled part for the semi-supervised learning approach). The parameters of the semi-supervised model were obtained by using the EM algorithm previously described. The fitted model was then used to predict the responses in the part that we set aside as the test set. These steps were repeated by selecting different part to set aside as the test set. All features used in the models are illustrated based on the example below as shown in Table 1.

**Example of a triple ($Q_p$, $Q_t$, $A_t$)**

**Prospective question**: anxiety medication for drug/alcohol addiction?

**Training question**: Is chlordiazepoxide/librium a good medication for alcohol withdrawal and the associated anxiety?

**Training answer:** chlordiazepoxide has been the standard drug used for rapid alcohol detox for decades and has stood the test of time. the key word is rapid the drug should really only be given for around a week. starting at 100 mg on day one and reducing the dose every day to reach zero on day 8. in my experience it deals well with both the physical and mental symptoms of withdrawal. looking ahead he will still need an alternative management for his anxiety to replace the alcohol. therapy may help, possibly in a group setting

**Table 1: List of features used in the model.**

| Type of Features | Features | Value |
|---|---|---|
| General Features | 1. Text length of $Q_p$ | 5 |
| | 2. Text length of $Q_t$ | 12 |
| | 3. Number of stop words contained in $Q_p$ | 1 |
| | 4. Number of stop words contained in $Q_t$ | 5 |
| | 5. $VS(Q_p, Q_t)$ | 3.7052 |
| | 6. The difference between $VS(Q_p, A_t)$ and $VS(Q_t, A_t)$ | 0.4303 |
| | 7. $DTW(Q_p, Q_t)$ | 29 |
| | 8. The difference between $DTW(Q_p, A_t)$ and $DTW(Q_t, A_t)$ | 14.5 |
| UMLS-based Features | 9. Number of overlapping words in $S_P$ and $S_T$ | 3 |
| | 10. Number of overlapping words in $S_P$ and $S_A$ | 3 |
| | 11. Binary variable indicating whether a set of overlapping words in ($S_P$, $S_T$) and ($S_P$, $S_A$) are different | 0 |
| | 12. Cardinality of the set difference of $S_P$ and $S_T$ | 4 |
| | 13. Cardinality of the set difference of $S_P$ and $S_A$ | 5 |

Sets $S_P$, $S_T$ and $S_A$ are sets of terms corresponding to UMLS concepts occurred in $Q_p$, $Q_t$ and $A_t$, respectively. General features are taken from previous work [7] while we introduce UMLS-based features into the model. Features 9 and 10 are calculated by counting the number of words contained in both sets. In order to obtain features 12 and 13, we find the elements that are in only one of the two sets.

Table 2 depicts examples of annotations in the corpus. The inter-rater agreement for random instances (10% of total) assigned to two independent reviewers is very good (95% confidence interval of kappa from 0.69 to 0.93). The procedure to identify an answer to a newly posted question is illustrated in Figure 2 after the usual split of the corpus in train and test.

**Table 2: corpus annotation examples**

| A target question | A training question | A training answer | Label |
|---|---|---|---|
| can fully recovered alcoholics drink again | can a recovered alcoholic drink again? | what they say at aa is that there is no such thing as permanent recovery from alcoholism. there are alcoholics who never drink again, but never alcoholics who stop being alcoholics….. | valid |
| can fully recovered alcoholics drink again | if both my parents are recovered alcoholics, will i have a problem with alcohol? | yes, there is a good chance that you could inherit a tendency towards alcoholism…. | invalid |
| anxiety medication for drug/alcohol addiction? | Is chlordiazepoxide/librium a good medication for alcohol withdrawal and the associated anxiety? | chlordiazepoxide has been the standard drug used for rapid alcohol detox for decades and has stood the test of time…. | valid |
| anxiety medication for drug/alcohol addiction? | negative affects of alcohol and adhd medication? | drinking in moderation is wise for everyone, but it is imperative for adults with adhd… | invalid |

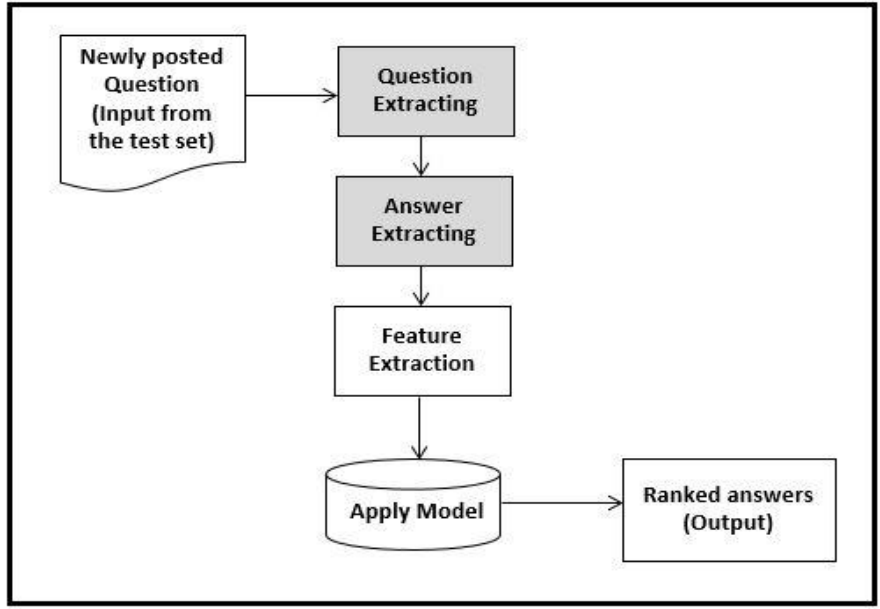

**Figure 2: Process flow of the testing step.**

The following evaluation metrics is used to test the overall performance of our algorithm.

1. Question-based evaluation metrics

- For this paper, we define "Overall Accuracy" as ratio of the number of questions with at least one "correct" answer divided by total number of questions in the test set. A test question is labeled as "correct" if our algorithm predicts at least one valid triple correctly. For the case that there is no valid answer in the question from the gold standard, we label it as "correct" if our algorithm predicts corresponding triplets as invalid.

- The Mean Reciprocal Rank (MRR) with a set of test questions Q is defined as

$$\text{MRR} = \frac{1}{|Q|} \sum_{i=1}^{|Q|} \frac{1}{\text{rank}_i}$$

where $rank_i$ is the position of a valid instance in manually sorted probabilities from the model. If there are more than one valid instance in any question, minimum value of $rank_i$ is used.

2. Triple-based evaluation metrics

Precision, recall, and the F1-Score can be used as standard measures for binary classification. We do not measure accuracy and ROC curves since the data set is heavily imbalanced.

## RESULTS

To test the algorithm, we obtained a total of 4,216 alcoholism-related QA threads from Yahoo! Answers. The sample outputs from our algorithm are shown in Figure 3, which indicates how our system could potentially be used by online advice seekers. In order to extract initial candidate answers in the rule-based answer extraction, our algorithm returns 8 instances for each prospective question (obtained from 2 different similarity measures where we extract at least 2 closest questions for each measure with 2 answers for each question). An example of output reported from the rule-based answer extraction is depicted in Figure 4.

| Input (a newly posted question) | Output (a possible answer) |
|---|---|
| **Question title:** Is there anything for alcohol withdrawal, medication recovery<br>**Question description:** Just like how there is some medication for drug addicts, like Suboxone, Methadone... etc. I wonder if theres anything like that for alcoholic? ..weed? lol.. or is there? - just wondering, thanks. | chlordiazepoxide has been the standard drug used for rapid alcohol detox for decades and has stood the test of time. the key word is rapid the drug should really only be given for around a week. starting at 100 mg on day one and reducing the dose every day to reach zero on day 8. in my experience it deals well with both the physical and mental symptoms of withdrawal. looking ahead he will still need an alternative management for his anxiety to replace the alcohol. therapy may help, possibly in a group setting. |
| **Question title:** Symtons off alcohol abuse<br>**Question description:** what side affecs can i expect when in stop drinking | i hope your meaning long term? not if you drink one night then suddenly stop all these return to normal. because they reactions will be piss poor when your pissed. fitness: will be down, the same as when you eat unhealthy etc.. you need to exercise to improve this again to a good standard. liver: if you havent abused alcohol for long & your still young, your liver can return to health naturally.. if it has been abused for years, you may have some liver disease, where it cant recoved.. reflexes: might be slightly worse of.. but shouldnt be affected too much, after normal exercise etc, should be back to normal. |
| **Question title:** Alcohol effects baby<br>**Question description:** So I'm four months preggers and evryones been telling me tht i shud not drink becuz its bad for the baby. i know pregnant women shudnt drink but thts just becuz they might do something they might not normally do thts bad, lik sleep with a random person or drive and crash. but wat if i drink and my friend watches over me and makes sure i don't do anything bad? Or shud i just not risk it? | no! if you suspect youre pregnant, dont go anywhere near alcohol -period. youre going to ruin the life of an innocent child who deserves more if you do. |

**Figure 3: System output.**

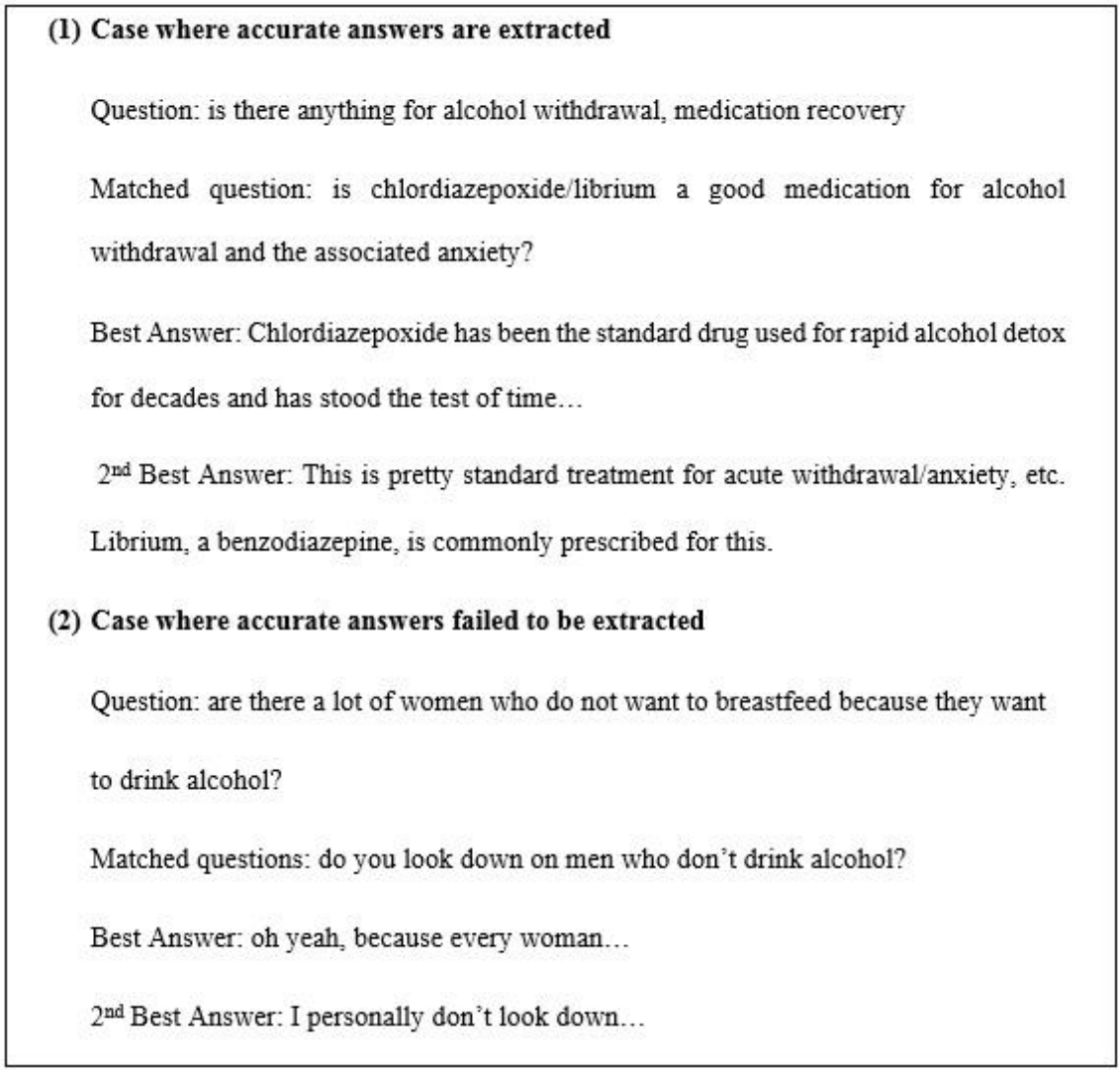

**Figure 4: An example result returned from the algorithm to determine candidate answers.**

A randomly selected set of 220 threads were manually annotated and used as labeled questions. Overall, 119 out of 220 questions, or 54.1 percent, have valid answers among those extracted in the rule-based answer extraction phase. After retrieving candidate answers, we further aim to re-rank them and select the best answer (if there is a valid answer). Note that each question corresponds to several candidate answers and thus multiple triplets ($Q_p$, $Q_t$, $A_t$). If at least one triplet is labeled as "valid," the corresponding question is also labeled as "valid."

Specifically, the semi-supervised learning model (EM) was trained on 1,553 labeled triplets (corresponding to 220 manually labeled questions) and 10,000 unlabeled triplets. In the training data of 1,553 labeled triplets, 297 triplets were manually labeled as "valid" and 1256 as "invalid." The typical 10-fold cross validation was implemented in order to validate the model.

We included all features listed in Table 1 in the models. In order to indicate a significance of each feature, we analyzed the feature set by using information gain. The information gain is based on the entropy function which is closely related with the objective function of the neural network NNET and logistic regression classifiers. The most influential features are the number of stop words contained in $Q_p$, the text length, the distance of $(Q_p, Q_t)$, as well as the number of overlapping UMLS words between $Q_p$ and $Q_t$, i.e. in $S_P$ and $S_T$. All information gains for these significant features are listed in Table 3.

**Table 3: Information gain score of 5 significant features**

| Features | Information gain |
| --- | --- |
| 1. Number of stop words contained in $Q_p$ | 0.0912 |
| 2. Text length of $Q_p$ | 0.0804 |
| 3. DTW$(Q_p, Q_t)$ | 0.0395 |
| 4. Number of overlapping words in $S_P$ and $S_T$ | 0.0393 |
| 5. VS$(Q_p, Q_t)$ | 0.0350 |

The best model was selected by varying the cutoff probability of being valid or invalid to obtain the maximum F1-score. We selected NNET, NNET_L2, SVM, and LOG approaches to train the model on a subset. For the SVM classifier, the probability was obtained by fitting a logistic distribution using maximum likelihood to the decision values provided by SVM.

The semi-supervised learning (EM) algorithm with 1 iteration trained with NNET_L2 gave the best performance for MRR and F1-score with a reasonable value of Overall accuracy while NNET performs best for Overall accuracy, as listed in Table 4. Each value in the table is the average across 100 different runs based on different random numbers in the algorithms and the test/train splits (details provided below). In Table 4, the numbers in bold represent the best value among different models and classifiers for each evaluation metric. The confusion matrices for 1 iteration of EM trained with 4 different classification models are provided in Figure 5.

**Table 4: Evaluation metrics**

| Evaluation Metrics | Supervised learning | | | | Semi-supervised learning (EM) | | | | | | | |
|---|---|---|---|---|---|---|---|---|---|---|---|---|
| | | | | | 1 iteration | | | | 10 iterations | | | |
| | NNET | NNET_L2 | SVM | LOG | NNET | NNET_L2 | SVM | LOG | NNET | NNET_L2 | SVM | LOG |
| Overall Accuracy | 0.5818 | 0.6993 | 0.6305 | 0.6245 | **0.8623** | 0.7105 | 0.6774 | 0.6473 | 0.8491 | 0.71 | 0.6783 | 0.6478 |
| MRR | 0.4216 | 0.5534 | 0.6224 | 0.6336 | 0.5686 | **0.6339** | 0.631 | 0.6266 | 0.5681 | 0.6332 | 0.6313 | 0.628 |
| F1-score | 0.1 | 0.3786 | 0.3045 | 0.3214 | 0.3222 | **0.3996** | 0.3667 | 0.3622 | 0.316 | 0.3977 | 0.3656 | 0.3626 |
| Precision | 0.0746 | 0.3614 | 0.4803 | 0.5073 | 0.2294 | 0.3981 | 0.4493 | 0.4421 | 0.2219 | 0.3942 | 0.4478 | 0.44 |
| Recall | 0.1433 | 0.4 | 0.241 | 0.2659 | 0.6801 | 0.4214 | 0.3239 | 0.3224 | 0.6562 | 0.4209 | 0.3229 | 0.3233 |

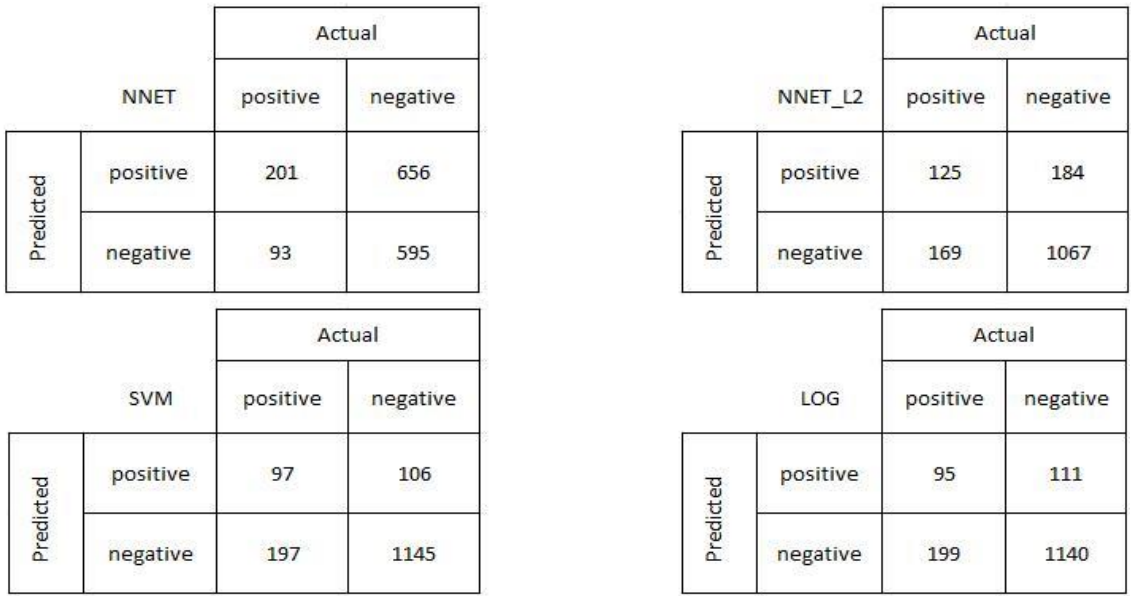

**Figure 5: The confusion matrices for 1 iteration of EM trained with NNET, NNET_L2, SVM, and LOG**

We performed 2 types of statistical hypothesis tests (t-tests) at the 0.05 level (95% confidence interval) to determine if two sets of evaluation metrics among the F1-score, overall accuracy, and MRR, obtained from different settings are significantly different from each other. First, randomness occurs within an algorithm such as the randomness in the stochastic gradient approach. Second, we consider randomness of assigning the test set, i.e. the training and test sets in 10-fold cross validation are randomly assigned. We performed both types of the hypothesis tests for all possible comparisons including the model implemented (pure classification vs semi-supervised), and among the 4 different classifiers based on the numbers reported in Table 4. Overall, the semi-supervised learning model is statistically significantly better than the corresponding supervised version for all evaluation metrics. This conclusion holds for both tests. Comparing between 1 and 10 EM iterations, the evaluation metrics are not statistically different from each other. This implies that the model parameters tuned by the EM algorithm are very close to the optimal values within 1 iteration.

We are also interested in understanding whether UMLS-based features (feature 9-13 listed in Table 1) play a role in predicting the validity of a candidate answer. Hence, we trained another model, which excludes all UMLS-based features, and compared the results (obtained from 1 iterations of EM trained with NNET_L2) with those from the original model as illustrated in Figure 6. The statistical tests at the 0.05 level showed significantly difference between the 2 models (with vs without UMLS-based features) for the 3 evaluation metrics. With UMLS-based features, the model gave a better performance, which is consistent across all evaluation metrics. This implies that these features played a role in distinguishing between valid and invalid answers.

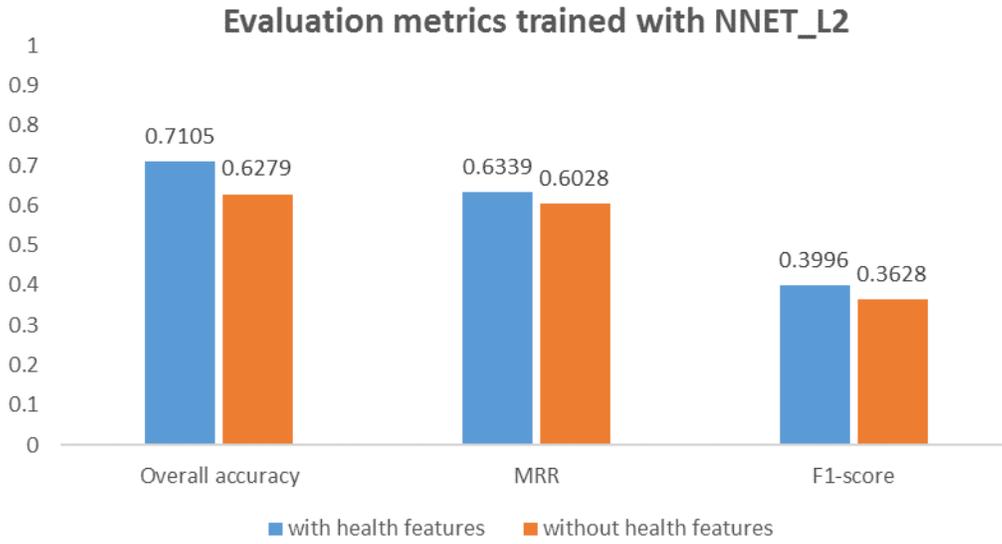

**Figure 6: Performance between the original and adjusted model to test significance of UMLS-based features (health features)**

## DISCUSSION

In this paper, we developed an automated QA system by using previously resolved QA pairs from a CQA site and evaluated it. Even though we used Yahoo! Answers as a data source, our algorithm can be adapted and applied to other CQA sites, in particular those related to healthcare where UMLS applies. Among different models and classifiers experimented, EM semi-supervised learning is better than pure supervised learning and 1 iteration of EM generally performs better than other models. Specifically, 1 iteration of EM with NNET gives the best performance in term of accuracy. NNET_L2 with 1 iteration of EM performs best in terms of the MRR and F1-scores. The NNET_L2 with 1 EM iteration is recommended to be used based on the case study data. Overall, the best model achieves an 86.2 percent accuracy and a 0.4 F1-score, which are significant given that the problem is challenging and the data is imperfect. Internet users typically provide responses in an ill-formed fashion. Our data also consists of a significant number of complex questions, e.g. a user discusses about his or her situation in 10-20 sentences and then asks whether s/he is an alcoholic. Moreover, some questions are very detailed; for example, the percentage of alcohol resulting from a given combination of chemical components. There is a trade-off between precision and recall. Some of these values listed in Table 4 are small as we aim to find a good balance between the two values. We intentionally maximize the F1-score which is a representative of both values. Precision and recall are reported in Table 4 for completeness. A comparison between the rule-based approach in the first phrase and the semi-supervised learning model in the second phrase reveals a significant improvement. The semi-supervised approach improves the accuracy of the model by 30% (approximately from 55% to 86%).

Comparing with Luo et al. [26] who retrieved the similar questions based on the distance measure, we relied on this idea with different approaches. In order to compute the similarity score between questions, we employed the DTW measure instead of relying on the vector-based distance measure. Luo et al. matching questions with information in data sources that are written and reviewed by experts; we strictly use only data from Yahoo! Answers, which are very noisy. For this reason the syntactic features proposed by Luo et al. might not be useful in our model. Unfortunately, not all libraries used in Luo et al.'s implementation are publicly available and thus direct comparison of the accuracy is not possible.

Shtok et al. [7] used resolved QA pairs to reduce the rate of unanswered questions in Yahoo! Answers. The experiment in Shtok et al. was also tested with health related questions and the accuracy as measured by the F1-score was 0.32. Our method, which trained a semi-supervised learning model with a smaller amount of manually labeled data compared to a supervised learning model used in [7], resulted in 0.4 F1-score. A better performance might be because of several reasons. First, we categorized questions in a corpus into different groups based on question keywords. Instead of computing the distance between a test question and all other questions in the corpus, categorizing questions reduces the scope of questions an algorithm needs to search. As we categorize collected questions into different groups based on question keywords, latent topics and "wh" question matching features used in Shtok's are not valuable in our context. Second, our algorithm also used multiple features related to the UMLS medical topics in order to enhance the model's performance when applied within the health domain where the Shtok's system was designed for a more general usage. While Shtok et al. relied on cosine distance, the Euclidean distance performed better in our evaluation. Among distance measures used in our work, more valid answers can be correctly identified with the DTW-based approach than the vector similarity measure, which can be observed when manually annotating the output from the rule-based answer extraction. In addition, our algorithm extracted multiple candidate answers retrieved from two closest QA pairs for each distance metric and the two best answers for each question. In each QA pair, both the best and the second best answer were extracted compared to Shtok et al. where only one best answer was extracted. Finally, we implemented semi-supervised learning to gain benefits from unlabeled data while Shtok et al. only relied on a supervised learning model in the re-ranking phase.

Using a semi-supervised learning model that leverages unlabeled data is reasonable against other traditional supervised learning models because obtaining labeled data is very expensive and time-consuming in practice. Since the features of the machine-learning algorithm are not specific to alcoholism, our system should be applicable for other related topics. On the other hand, it would be possible to increase the accuracy for "alcoholism" if we use specific features such as concepts related to alcoholism.

In summary, the main novelty and advantages of our work against other works include the DTW-based distance approach, UMLS-based features, the semi-supervised learning algorithm, and the data set used in the study. We introduce a novel distance measures, the DTW-based approach which performs better than the typical vector-space distance method. UMLS-based features are included to enhance the model applied in the health care domain in addition to the general features in [7]. Our system is trained and tested only on the online information without any additional sources. Also, obtaining the annotation from online data can be very difficult and time-consuming. This stresses the significance of using semi-supervised learning rather than a typical supervised learning algorithm.

For the machine learning component, the distance between a test question and other questions in the training data set is important in distinguishing valid and invalid answers. The closer the distance is, the higher the chance of the corresponding answer being valid. Matching UMLS terms, which imply a closer similarity between questions, play a role in determining the validity of the answer. Even though UMLS-based features show lower information gain, the model with these features included is significantly better across all evaluation metrics. The overall accuracy is improved by 8% when these features are included.

Information gain shows that number of stop words contained in a test question and the underling text length are the best indicators for differentiating between valid and invalid answers. We note that the number of content-rich words, represented as text length minus the number of stop words, is also taken indirectly into account by these two features. We fitted the model without the number of stop words feature compared with the full model. Even though these two models are not statistically different, we include the number of stop words feature in the model as previously done in [7].

**Limitations and future work**

The main limitation of our work is the lack of assessment of the model's generalizability. Even though our algorithm is generic and does not include any features that are specific to the topic of alcoholism, we have not validated it in different domains as we do not have available data. Approximately 30 percent (obtained from a preliminary observation) of all questions cannot be answered based on existing answers; some of these questions also require additional resources that are more technical and reliable, such as medical textbooks, journals and guidelines.

**CONLUSIONS**

The question-answering system developed in this work achieves reasonably good performance in extracting and ranking answers to questions posted in CQA sites. Our work is a promising approach for automatically answering alcoholism-related questions obtained from CQA sites based only on past QA that is used as a case study. Also, our system can potentially be applied to other health-care domain questions asked by online healthcare communities. The system and the gold standard corpus are available in github [35].

**Acknowledgments**

We are grateful to Dr. Jina Huh from Michigan State University for providing the Yahoo! Answers dataset and Dr. Kalpana Raja from Northwestern University for helping with UMLS. This work was partly supported by funding from the National Library of Medicine grant R00LM011389.

**Conflicts of interest**

None declared.

**Abbreviations**

CQA: Community-based question answering
QA: Questions and answers
UMLS: Unified medical language system
PICO: Patient/population, intervention, comparison, outcome
EM: Expectation maximization
DTW: Dynamic time warping
NNET: Neural networks
SVM: Support vector machine
MRR: Mean reciprocal rank
ROC: Receiver operating characteristic